\documentclass[conference]{IEEEtran}
\IEEEoverridecommandlockouts
\usepackage{cite}
\usepackage{amsmath,amssymb,amsfonts}
\usepackage{algorithmic}
\usepackage{graphicx}
\usepackage{textcomp}
\usepackage{xcolor}

\def\BibTeX{{\rm B\kern-.05em{\sc i\kern-.025em b}\kern-.08em
    T\kern-.1667em\lower.7ex\hbox{E}\kern-.125emX}}
\begin{document}

\title{Resilience of Wireless Ad Hoc Federated Learning \\ against Model Poisoning Attacks 
}

\author{\textbf{Preprint version}\\Naoya Tezuka, Hideya Ochiai, Yuwei Sun, Hiroshi Esaki, \IEEEmembership{Member, IEEE},
\thanks{Naoya Tezuka, Hideya Ochiai, Yuwei Sun, and Hiroshi Esaki are with the Graduate School of Information Science and Technology, University of Tokyo, Tokyo, 113-8656, Japan (e-mail: ochiai@elab.ic.i.u-tokyo.ac.jp).}
\thanks{This work was accepted by IEEE International Conference on Trust, Privacy and Security in Intelligent Systems, and Applications (TPS-ISA), 2022.}
}


\maketitle

\begin{abstract}
Wireless ad hoc federated learning (WAFL) is a fully decentralized collaborative machine learning framework organized by opportunistically encountered mobile nodes. Compared to conventional federated learning, WAFL performs model training by weakly synchronizing the model parameters with others, and this shows great resilience to a poisoned model injected by an attacker. In this paper, we provide our theoretical analysis of the WAFL's resilience against model poisoning attacks, by formulating the force balance between the poisoned model and the legitimate model. According to our experiments, we confirmed that the nodes directly encountered the attacker has been somehow compromised to the poisoned model but other nodes have shown great resilience. More importantly, after the attacker has left the network, all the nodes have finally found stronger model parameters combined with the poisoned model. Most of the attack-experienced cases achieved higher accuracy than the no-attack-experienced cases.
\end{abstract}


\section{Introduction}

Cyber attacks on machine intelligence may result in unexpected catastrophic disasters especially when we fully rely on the machine's decisions. Recent advancement in federated learning has allowed training machines without exchanging privacy-sensitive data. Instead, they exchange the model itself, which means that the access (read and write) to the intelligence is open to any entities including malicious entities.

Because of the need for ad hoc federation among the nodes physically nearby, wireless ad hoc federated learning (WAFL) has been proposed \cite{ochiai2022wireless}. WAFL assumes that autonomous vehicles, smart devices, or sensor nodes have multi-vendor issues and cannot rely on a centralized parameter server \cite{konevcny2016federated,li2020federated,tan2022towards} operated by third-party organizations. Instead, WAFL utilizes ad hoc connections among those devices for collaborative learning in a fully distributed manner without any centralized or third-party mechanisms as cooperative intelligent transport systems do \cite{festag2014cooperative}. Here, security in such an open and fully distributed learning framework is an important topic.

In this paper, we study model poisoning attacks against legitimate nodes that are collaboratively learning with WAFL. We especially focus on analyzing the resilience of WAFL against model poisoning attacks. 

Compared to conventional federated learning, we consider that WAFL has more resilience because of its fully distributed nature. Federated learning has a critical attack point ``global model'', which is synchronized among the participating nodes. If the model is somehow poisoned by a malicious node, the compromised model would be synchronized to all the nodes. 

\begin{figure}
\centering
\includegraphics[width=0.40\textwidth]{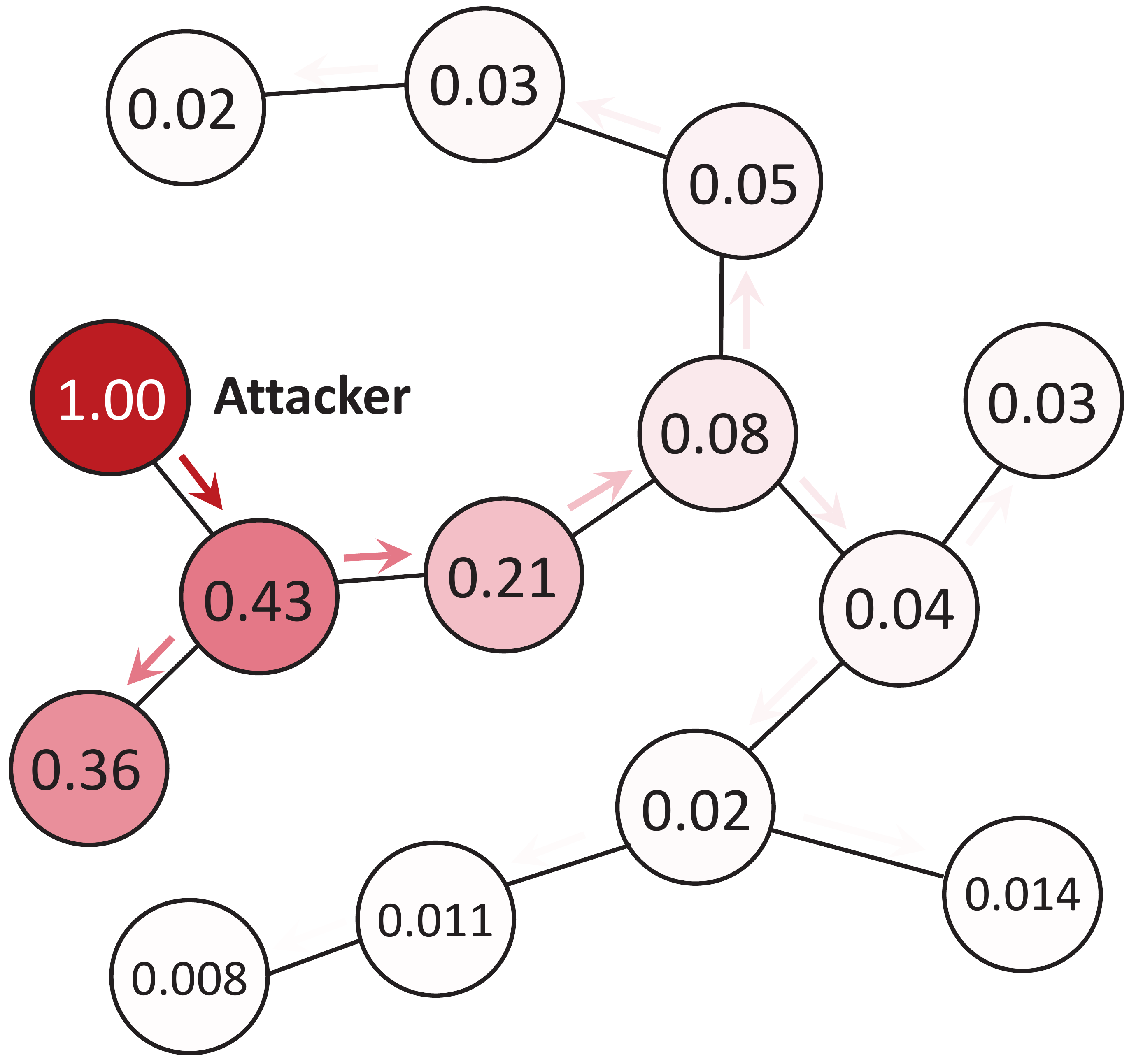}
\caption{Impact of model poisoning attack in wireless ad hoc federated learning. When an attacker injects a malicious model persistently into one of the nodes in a WAFL network, the impact (i.e., 1.00 in the figure) would be propagated in the network, but rapidly overrode by the training of the legitimate nodes showing its resilience as the hop count increases.}
\label{fig:impact}
\end{figure}

However, WAFL does not have such a critical attack point. In WAFL, each node has its own generalized model, developed by model exchange and aggregation among the nodes encountered. These generalized models are weakly-synchronized with each other at every contact, but not fully synchronized as a global model. Thus, we expect that an attack on one of the generalized models cannot fully poison all the models possessed by the distributed nodes in the network. Besides, because legitimate nodes continuously inject legitimate models for developing generalized models, the impact of the attack would be propagated but rapidly weakened as the hop count increases from the attacker (Fig. \ref{fig:impact}).

In this paper, we provide a theoretical analysis of the propagation of the poisoned model, also showing the flow of resilience by the legitimate node. Then, we carry out a benchmark-based evaluation with Non-IID MNIST dataset and a static line network topology (1) to observe the confusions of the attacked nodes, (2) to evaluate the accuracy drops, (3) to evaluate poisoned model propagation and the resilience.

The contributions of this work are summarized as follows:

\begin{itemize}
\item We provide a theoretical analysis of the poisoned model propagation in the WAFL network, showing the mechanism of resilience against the poisoning attack.
\item We define four attack classes as examples of model poisoning attacks.
\item This paper focuses on benchmark-based evaluation with MNIST and simple network topology for understanding the resilience of WAFL against four classes of model poisoning attacks.
\item We have discovered that attack-experienced WAFL networks can achieve higher accuracy than the no-attack case as well as showing their resilience to the attack.
\end{itemize}

This paper is organized as follows. In section II, we address the related works. In section III, we provide a theoretical analysis of the resilience of WAFL against model poisoning attacks. Section IV shows our evaluation study. After giving future research directions in section V, we conclude this paper in section VI.

\begin{figure*}
\centering
\includegraphics[width=0.90\textwidth]{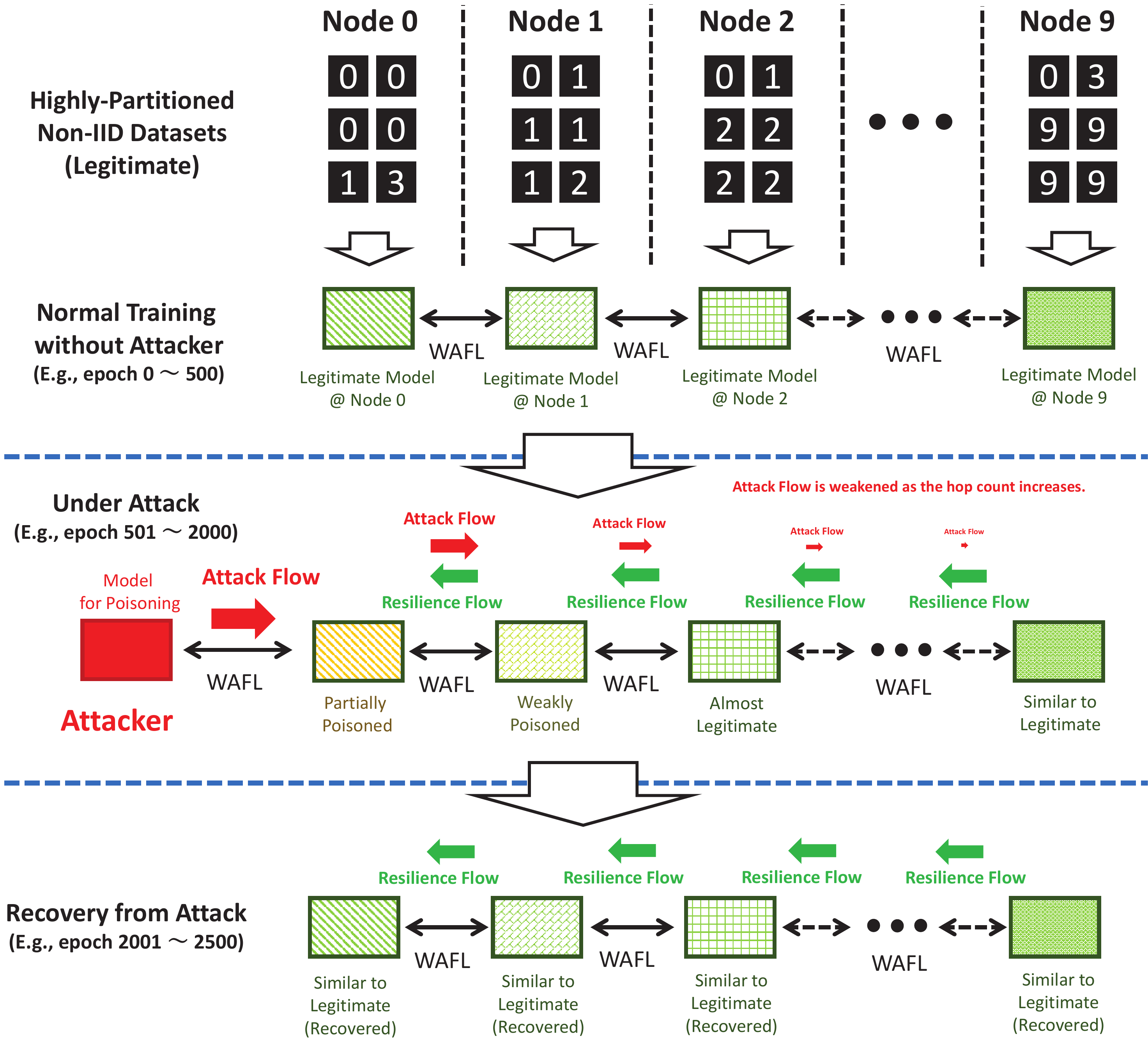}
\caption{Attack scenario during the training of WAFL. Nodes are training the model over highly-partitioned Non-IID datasets by exchanging and aggregating their models based on the WAFL algorithm. An attacker injects poisoned models into one of the nodes, expecting to disseminate it to all the nodes. However, WAFL shows resilience to such a poisoned model through its legitimate learning process. After the attacker is removed from the network, all the nodes will soon recover from the attack.
}
\label{fig:resilience_overview}
\end{figure*}

\section{Related Work}

Federated learning was proposed with expectations of protecting users' privacy by exposing the global model to participating users and local training without exchanging the user's data. However, this model exposure including the downloading/uploading interface brought another risk to the machine learning model such as poisoning attack\cite{fang2020local}, and privacy leakage\cite{nasr2019comprehensive,geiping2020inverting,sun2021information}. 

Depending on the goals of the attacker, many poisoned models have been studied \cite{fang2020local, sun2022semi, suciu2018does}. Backdoor attack \cite{li2022backdoor} has been mainly considered by poisoning the data while training, but is also studied in the context of federated learning \cite{bagdasaryan2020backdoor, wang2020attack}. The idea is to upload a backdoor-enabled model to a central parameter server, replacing the global model with it.

Some defense mechanisms for model poisoning attack in federated learning has been also studied \cite{ozdayi2021defending, cao2021provably} in the recent literature. However, all of these works assume a centralized parameter server and a global model for attacking and defending the models.

In peer-to-peer federated learning, Sun et. al. \cite{sun2020blockchain} and Li et. al.\cite{li2020blockchain} have proposed Blockchain-based protection mechanisms against such a malicious model update. They introduce a consensus mechanism for the acceptance of model updates. If the update could not get voted approval by more than half of the committees, it will not be aggregated into the global model. However, the work still assumes a single global model, which cannot be applicable to WAFL.

Wireless ad hoc federated learning (WAFL) seems to have the same risks as conventional federated learning in that WAFL also exposes model parameters among the participated nodes. However, because WAFL is totally distributed without any centralized mechanisms, model poisoning attacks made to a node cannot be easily propagated to all the participating nodes. As long as we surveyed, the analysis of model poisoning attacks to such a distributed framework was not studied in the past.

\section{Resilience against Model Poisoning Attack}

In this section, we analyze WAFL's distributed algorithm regarding poisoned model propagation and its resilience.

\subsection{WAFL overview}

We assume a set of nodes $N$ that collaboratively train models using the data they locally have. Let $\theta^{(n)}$ be model parameters of node $n \in N$, and $D^{(n)}=(X^{(n)},Y^{(n)})$ be a set of data samples at node $n$. With function $f$, the output of model $\hat{y}$ to an input $x \in X^{(n)}$ at node $n$, can be formulated as follows:

\begin{equation}
\hat{y}=f(x,\theta^{(n)})
\label{eq:model_prediction}
\end{equation}

With appropriate loss function, $\theta^{(n)}$ can be trained to fit specifically for $D^{(n)}$. WAFL assumes this self-training process before the collaboration (i.e., model exchange and aggregation) with other nodes.

In WAFL, nodes are expected to communicate with each other only when they are neighbors, i.e., within the radio range for communication. For describing this, let $nbr(n)$ be a set of neighbors of node $n$ at a certain time. Here, $nbr(n)$ does not include itself, i.e., $n \notin nbr(n)$. $nbr(n)$ may dynamically change based on the physical mobility of the nodes. Node $n$ can directly communicate with $nbr(n)$ over wireless channels such as Wi-Fi (ad hoc mode) or Bluetooth. For simplicity, we assume the communication links are symmetric: $\forall k, n:  k \in nbr(n) \Leftrightarrow n \in nbr(k)$.

In WAFL, node $n$ receives all the model parameters from $nbr(n)$: i.e., $\{ \theta^{(k)} \vert k \in nbr(n)\}$. Then, node $n$, aggregates the models with the local model $\theta^{(n)}$ by the following formula. 

\begin{equation}
\theta'^{(n)} = \theta^{(n)} + \lambda \frac{\sum_{k \in nbr(n)}{(\theta^{(k)} - \theta^{(n)})}}{\vert nbr(n) \vert +1}.
\label{eq:model_aggregation}
\end{equation}

Here, $\lambda$ is a coefficient parameter. The aggregated model $\theta'^{(n)}$ needs additional mini-batch-based training with its local data. This can be formulated as follows:

\begin{eqnarray}
\small
\theta'^{(n)}_{1}&=&\theta'^{(n)}_{0}-\eta\nabla J \left( \theta'^{(n)}_{0}, D^{(n)}_0 \right), \nonumber \\ 
\theta'^{(n)}_{2}&=&\theta'^{(n)}_{1}-\eta\nabla J \left( \theta'^{(n)}_{1}, D^{(n)}_1 \right), \nonumber \\ 
&\vdots& \nonumber \\
\theta''^{(n)}&=&\theta'^{(n)}_{b-1}-\eta\nabla J \left( \theta'^{(n)}_{b-1}, D^{(n)}_{b-1} \right).
\label{eq:minibatch_in_model_aggregation}
\end{eqnarray}

Here, the index number $i$ in $\theta'^{(n)}_{i}$ indicates the the mini-batch round $i \in [0, \ldots, b-1]$. $D^{(n)}_i \subset D^{(n)}$ is the dataset extracted for the mini-batch $i$. $J$ is the aggregated loss to the predictions $\hat{y}$ from the mini-batch training dataset $D^{(n)}_i$ under Eq. (\ref{eq:model_prediction}), and $\eta$ is the learning rate.  After finishing this mini-batch-based training, the node adopts $\theta''^{(n)}$ as the next epoch's initial model parameter:

\begin{equation}
\theta^{(n)} \leftarrow \theta''^{(n)}.
\end{equation}

Then, the node starts the next training epoch from Eq. (\ref{eq:model_aggregation}).

After interacting with many nodes encountered, $\theta^{(n)}$ becomes generalized and it will be able to correctly predict the data samples that node $n$ does not have but others have. Please note that during this process, WAFL does not rely on any centralized mechanism and achieves model training without exchanging the local data.

\subsection{Model Poisoning in WAFL}

Let's assume a malicious node $m$. We denote $\theta^{(m)}$ by the poisoned model that $m$ injects into the WAFL framework.

\begin{figure*}
\centering
\includegraphics[width=0.90\textwidth]{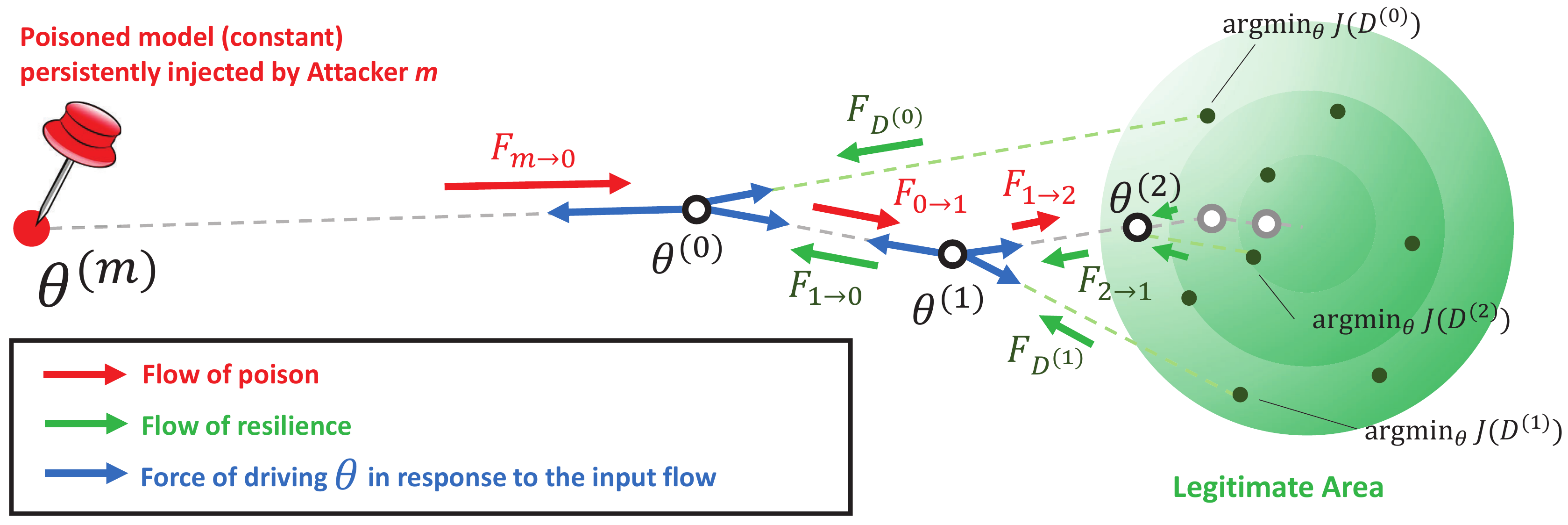}
\caption{The dynamics of model parameters under an attack in the parameter space. 
The flows of poisons are illustrated with red arrows, and the flows of legitimates are illustrated with green arrows. In response to the input flows, the model will be forced to move in the parameter space, which is illustrated by blue arrows. In the balanced situation of the forces, $\theta^{(0)}$ and $\theta^{(1)}$ are pulled out of the legitimate area, but as the hop count increases from the attacker, they can be seen as legitimate because those models are also pulled toward the legitimate area and remain there. Regarding the argmins in the legitimate area, please refer to \cite{ochiai2022wireless}.}
\label{fig:parameter_space}
\end{figure*}

In the case of the well-known model replacement attack in conventional federated learning\cite{bagdasaryan2020backdoor}, $\theta^{(m)}$ will be amplified before uploading to the central parameter server, but we do not take this approach because such amplification can be easily detected by checking the differences of model parameters. Besides, the attacker does not know the important parameter for precise model replacement, i.e., the number of neighbor nodes that the target node has, because of the hidden node problem \cite{rahman2006hidden, ray2005evaluation}. So, in this paper, we consider, more general, model poisoning attacks rather than the specific model replacement attack. 

Because of the physically-distributed nature of the WAFL environment, malicious node $m$ can inject poisoned model $\theta^{(m)}$ into a few nodes -- not all the nodes at the same time. It can inject directly into the nodes in $nbr(m)$, but others are not accessible from $m$.

However, the attacker can expect the attacked node to disseminate the poisoned model somehow to the entire WAFL network based on the connections of the nodes.



\subsection{Propagation of Poisoned Model and Resilience}

Fig. \ref{fig:resilience_overview} shows the propagation of the poisoned model on a WAFL network injected by an attacker.

In this scenario, we assume that legitimate nodes 0, \ldots, and 9 collaboratively make WAFL's training up to 500 epochs, and then the attacker mounts to this network at node 0. The attacker persistently injects its poisoned model via node 0, and it disseminates to the entire network based on the WAFL's aggregation function, i.e., Eq. (\ref{eq:model_aggregation}). After a while at epoch 2000 in the figure, the attacker is removed from the WAFL network, and the whole network goes into the recovery process.

In order to understand the dynamics of WAFL with a simpler configuration, let us consider the static\_line topology presented in \cite{ochiai2022wireless}. In this network topology, nodes 0 to 9 are sequentially and statically connected in a cascaded manner. If attacker $m$ mounts this network at node 0, the topology becomes $m-0-1-\ldots-9$ as Fig. \ref{fig:resilience_overview}.

We consider node $c \in [0, \ldots, 8]$, and nodes $l$ and $r$ so that $l$ is the left side of node $c$ and $r$ is the right side of node $c$. Here, $c$ stands for the central. Then, Eq. (\ref{eq:model_aggregation}) can be converted to,

\begin{equation}
\theta'^{(c)} - \theta^{(c)} = F_{l \rightarrow c} + F_{r \rightarrow c}.
\label{eq:model_propagation}
\end{equation}

This means that the update of $\theta^{(c)}$ by the aggregation is the summary of $F_{l \rightarrow c}$ and $F_{r \rightarrow c}$. These are the flows of model parameters from nodes $l$ to $c$ and $r$ to $c$ respectively, which we defined as:
\begin{eqnarray}
F_{l \rightarrow c}&=&\lambda \frac{\theta^{(l)} - \theta^{(c)}}{3}, \\ 
F_{r \rightarrow c}&=&\lambda \frac{\theta^{(r)} - \theta^{(c)}}{3}.
\end{eqnarray}

Practically, we specify coefficient $\lambda \in (0, 1]$. In the case of $\lambda=1$, one-third of the differences will be added to $\theta^{(c)}$ in a single aggregation epoch. This means that the same amount is subtracted from the model of the other node.

\begin{table*}
\centering
\caption{The configuration of Non-IID MNIST data samples. 90\% of node $n$'s data is composed of label $n$'s samples, and the other 10\% is uniformly composed of the other label's samples without any overlaps among the nodes. \label{tab:noniid_distribution}}
\begin{tabular}{c|cccccccccc|c}
\hline
Node&L0&L1&L2&L3&L4&L5&L6&L7&L8&L9&Summary \\ \hline
0&5341&76&64&76&61&57&57&61&74&50&5917 \\
1&79&6078&67&52&57&59&58&79&59&68&6656 \\
2&58&67&5374&80&64&68&87&73&57&61&5989 \\
3&68&74&73&5537&51&56&72&65&67&77&6140 \\
4&73&67&80&67&5301&59&53&69&70&64&5903 \\
5&60&66&57&74&68&4896&61&59&66&69&5476 \\
6&52&78&53&56&58&66&5312&56&65&65&5861 \\
7&67&90&66&74&55&61&65&5683&63&77&6301 \\
8&59&80&59&54&63&48&88&67&5268&57&5843 \\
9&66&66&65&61&64&51&65&53&62&5361&5914 \\ \hline
Summary&5923&6742&5958&6131&5842&5421&5918&6265&5851&5949&60000 \\ \hline
\end{tabular}
\end{table*}

Without the local mini-batches of Eq. (3), if attacker $m$ persistently injects a constant model $\theta^{(m)}$ to node 0, $\theta^{(0)}$ will be converged to $\theta^{(m)}$, and other models $\theta^{(1)}$, $\theta^{(1)}$, \ldots, $\theta^{(9)}$ will be also converged to $\theta^{(m)}$. This can be also driven by the fact that Eq. (2) is a diffusion equation that allows the propagation of model parameters to the entire network.

However, WAFL has the process of the local mini-batches and this will show resilience to the propagation of the poisoned model. To understand this, we define $F_{D^{(n)}}$ as the flow of model parameters injected to $\theta^{(n)}$ by local mini-batches on $D^{(n)}$,

\begin{equation}
F_{D^{(n)}} = \theta''^{(n)} - \theta'^{(n)}.
\end{equation}

Then, we get the following model updates,

\begin{equation}
\theta^{(c)}_{e+1}-\theta^{(c)}_e= F_{l \rightarrow c} + F_{r \rightarrow c} + F_{D^{(c)}}.
\end{equation}

Here, the number $e$ in $\theta^{(c)}_e$ shows the epoch number of the model - not the mini-batch number.

Under the model poisoning attack, if $\theta^{(m)}$ is constant, we can expect that $\theta^{(c)}_{e+1}-\theta^{(c)}_e$ will converge to 0. Thus, we will get the following law under the convergence,

\begin{equation}
F_{l \rightarrow c} + F_{r \rightarrow c} + F_{D^{(c)}}=0.
\end{equation}

Fig. \ref{fig:parameter_space} shows the parameter space of the converged scenario, where all the input flows are balanced at all the nodes, except at the attacker. In this figure, the flows of poisons are illustrated with red arrows, and the flows of legitimates are illustrated with green arrows. In response to the input flows, the model will be forced to move in the parameter space, which is illustrated by blue arrows. In the balanced situation of the forces, in the figure's example, $\theta^{(0)}$ and $\theta^{(1)}$ are pulled out of the legitimate area, but as the hop count increases from the attacker, they can be seen as legitimate because those models are also pulled toward the legitimate area and remain there.

\begin{figure}
\centering
\includegraphics[width=0.45\textwidth]{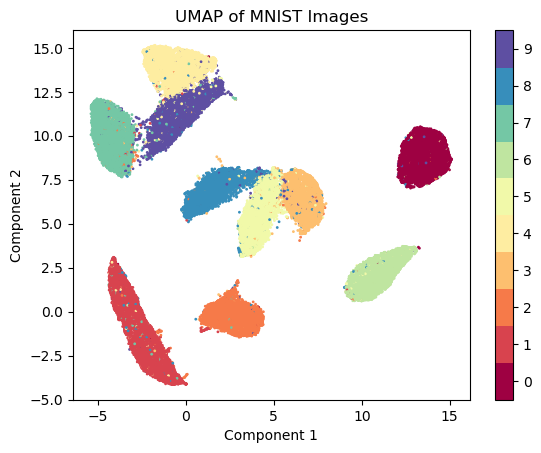}\\
\caption{UMAP projections of the MNIST training dataset. Classes 4 and 9 are similar, whereas classes 1 and 6 are dissimilar.}
\label{fig:MNIST_latent_space}
\end{figure}

\subsection{Attack Classes}

In this paper, we define four attack types for model poisoning attacks.

\subsubsection{Random Model Parameter (RMP)}

RMP attack is a poisoning attack that injects a model with constant random parameters. Such a poisoned model provides poor outputs for any inputs. The mixture with this model is expected to generate confusion in all the classes at legitimate nodes.

\subsubsection{Swap Similar Classes (SSC)}

By swapping the class labels, for example, 4 and 9, an attacker can generate a poisoned model that gives confusion to the outputs of those specific classes. In this example, this model will output 9 for the inputs of true 4 samples. The effect of the class swapping can be different depending on the similarities of the classes. Swap similar classes (SSC) attack is an attack of model injection trained by swapping a similar class pair. Examples of similar pairs in the MNIST dataset are, (4,9), (3,8), (3,5), and (7,9). Those similar pairs are located closely in the feature space in Fig. \ref{fig:MNIST_latent_space}.

\subsubsection{Swap Dissimilar Classes (SDC)}

Another class swapping option is to select a dissimilar pair of classes. For example, in the MNIST dataset, the pairs of (1,6), (0,4), (0,9), and (0,7) are dissimilar. We define the model injection generated by swapping labels of a dissimilar pair as swap dissimilar classes (SDC) attack.

\subsubsection{Class Output Depression (COD)}

By depressing the output of a specific class, the attacker can generate a poisoned model that does not correctly predict the input of specific class samples. For example, by overriding the true label of 0 by other labels (such as specifically 6 or other random labels), the attacker can generate a dataset that does not contain 0 as a label but the representations of the real samples are 0. By training with this overridden dataset, the attacker can generate a poisoned model that depresses such targeted class output.

\begin{figure*}
\centering
\includegraphics[width=0.98\textwidth]{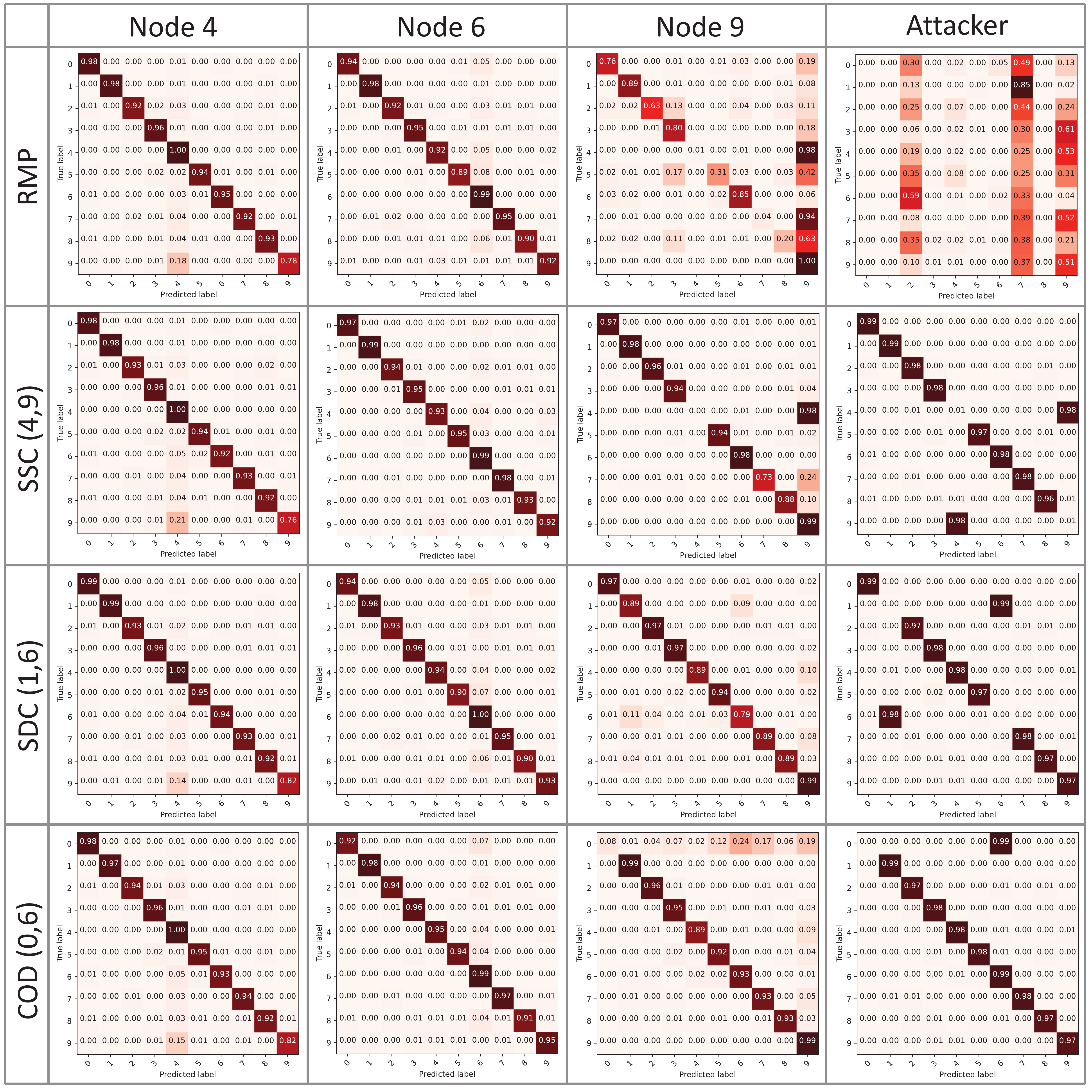}
\caption{Confusion matrices under the attack at epoch 2000. The attacker is mounted on Node 9, persistently injecting the poisoned models from epoch 500.} 
\label{fig:ConfusionMatrix}
\end{figure*}

\begin{figure*}
\centering

  \begin{minipage}[b]{0.32\textwidth}
  \raggedright
  \includegraphics[keepaspectratio, scale=0.38]{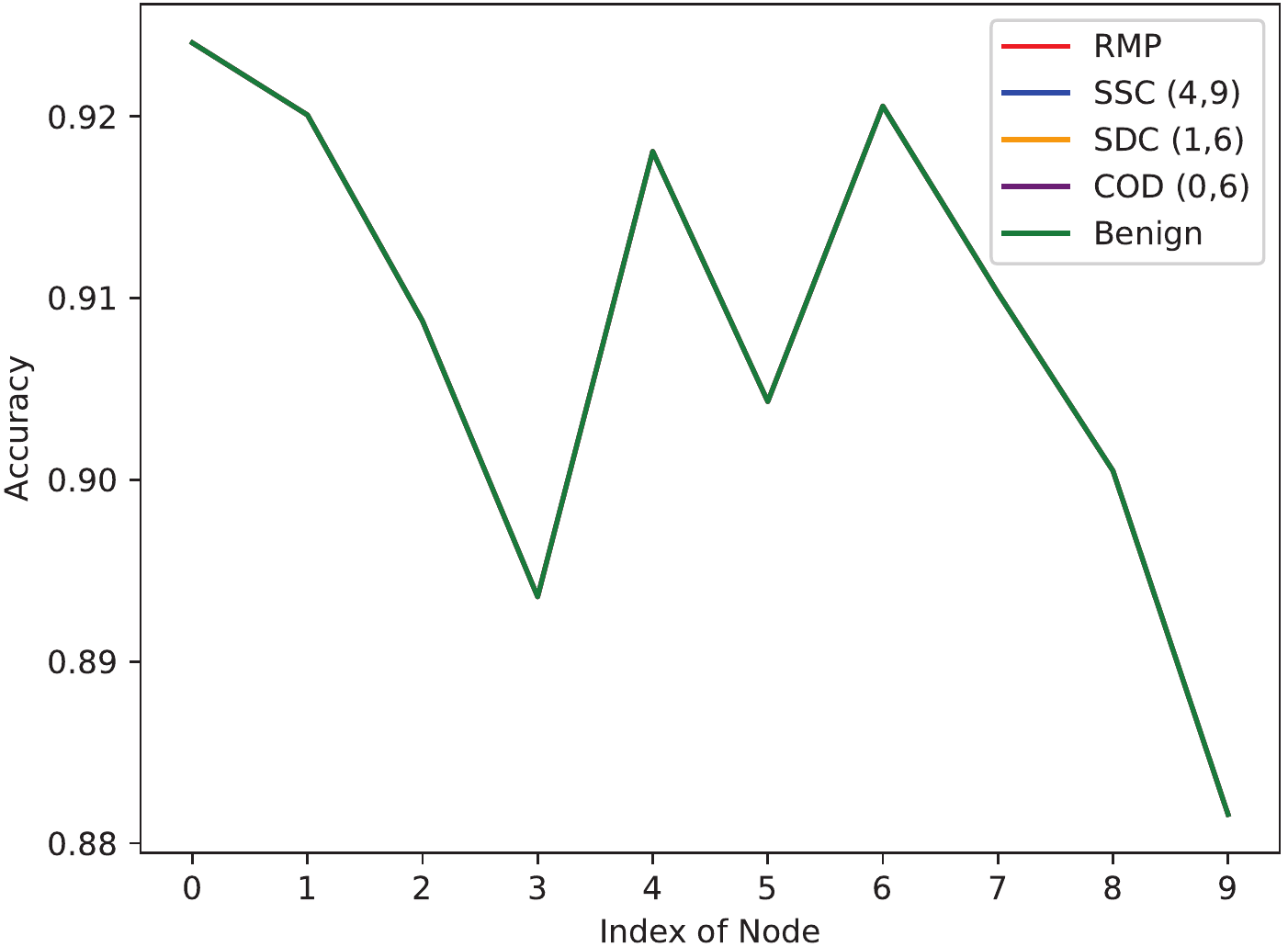}\\
  \scriptsize{\hspace{6em}(a) Before the Attack on Node 0}
  \end{minipage}
  \begin{minipage}[b]{0.32\textwidth}
  \raggedright
  \includegraphics[keepaspectratio, scale=0.38]{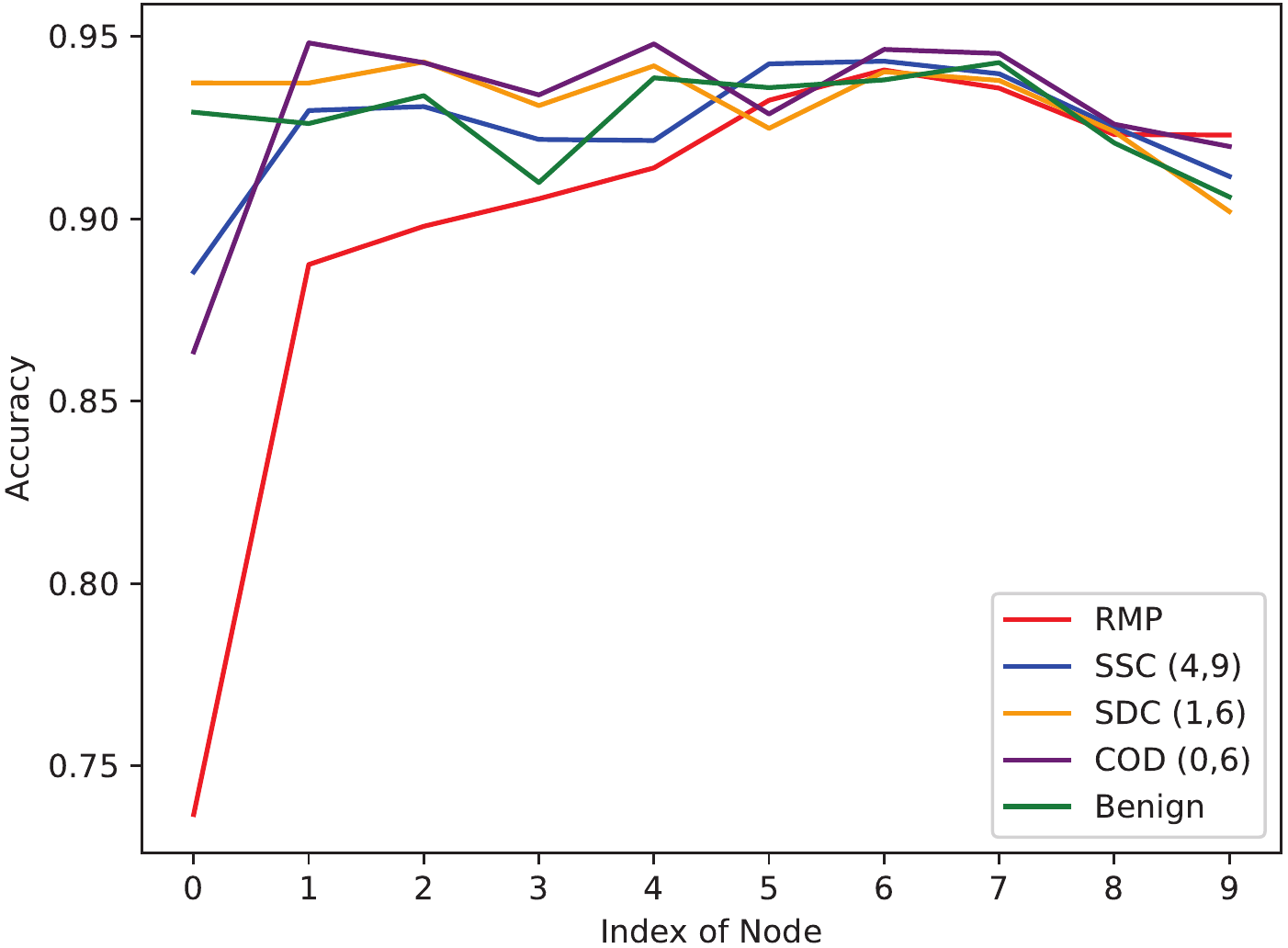}\\
  \scriptsize{\hspace{5em}(b) Under the Attack on Node 0}
  \end{minipage}
  \begin{minipage}[b]{0.32\textwidth}
  \raggedright
  \includegraphics[keepaspectratio, scale=0.38]{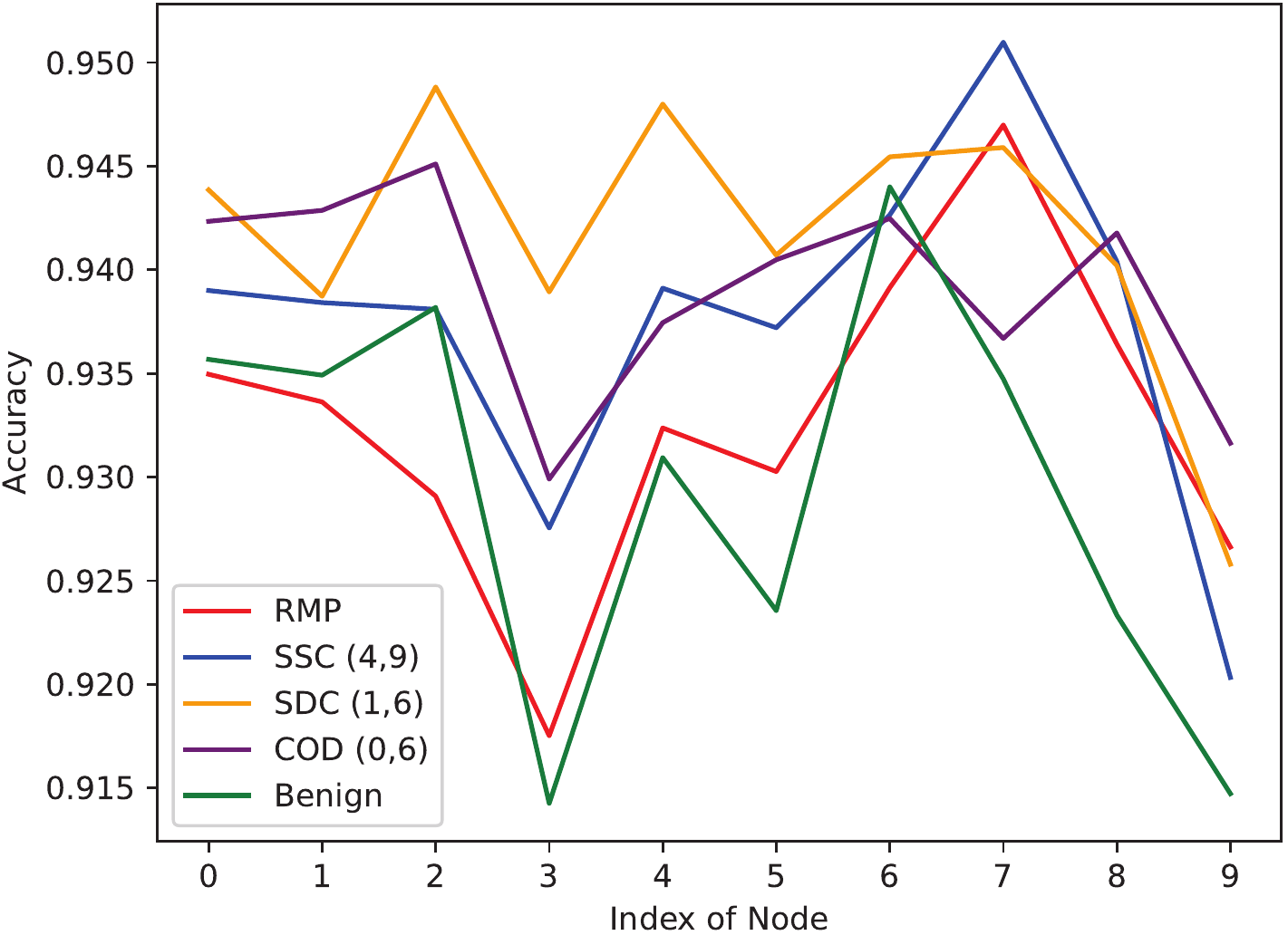}\\
  \scriptsize{\hspace{5em}(c) After the Attack on Node 0}
  \end{minipage}

  \vspace{1em}
  
  \begin{minipage}[b]{0.32\textwidth}
  \raggedright
  \includegraphics[keepaspectratio, scale=0.38]{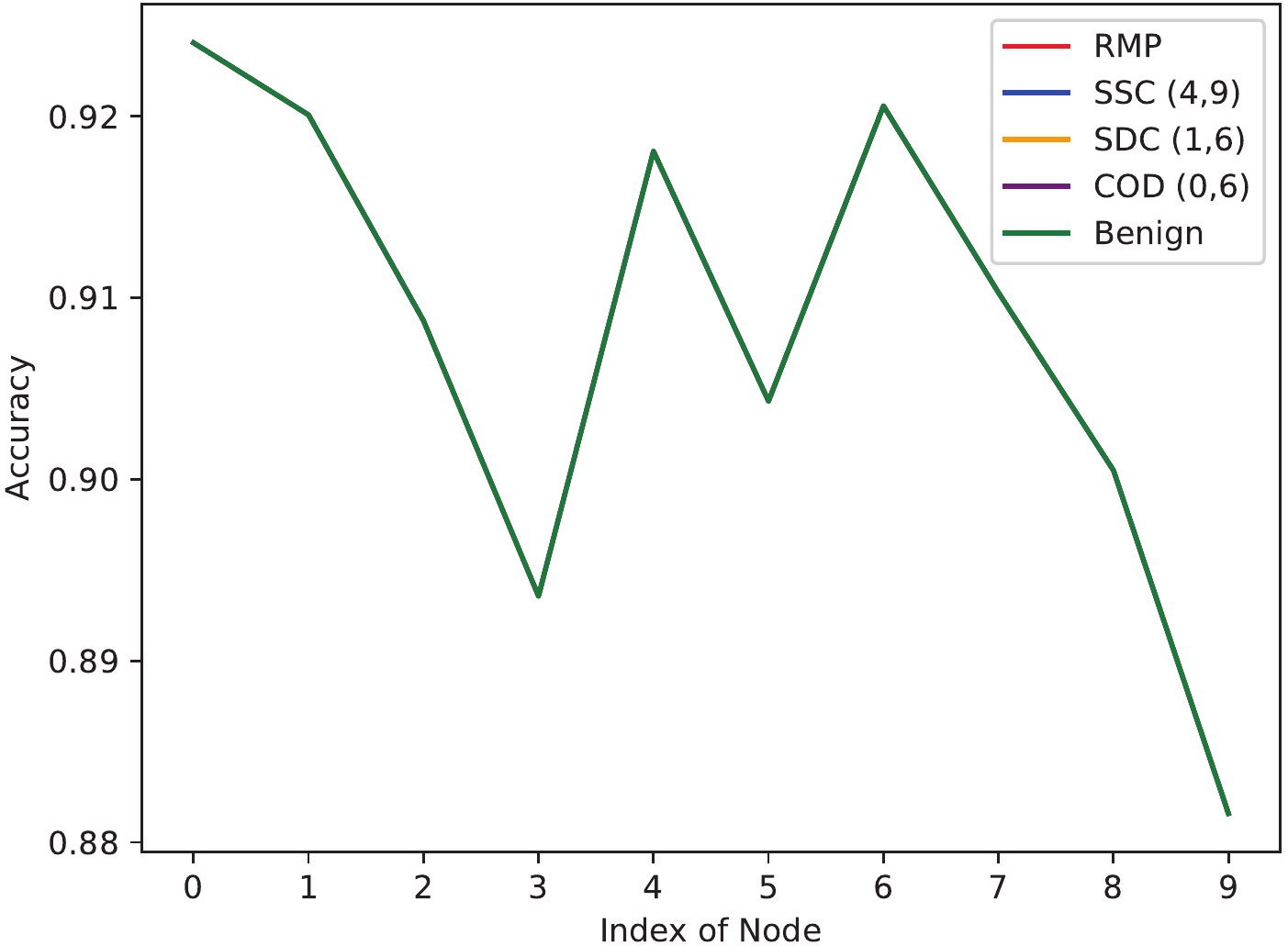}\\
  \scriptsize{\hspace{6em}(d) Before the Attack on Node 9}
  \end{minipage}
  \begin{minipage}[b]{0.32\textwidth}
  \raggedright
  \includegraphics[keepaspectratio, scale=0.38]{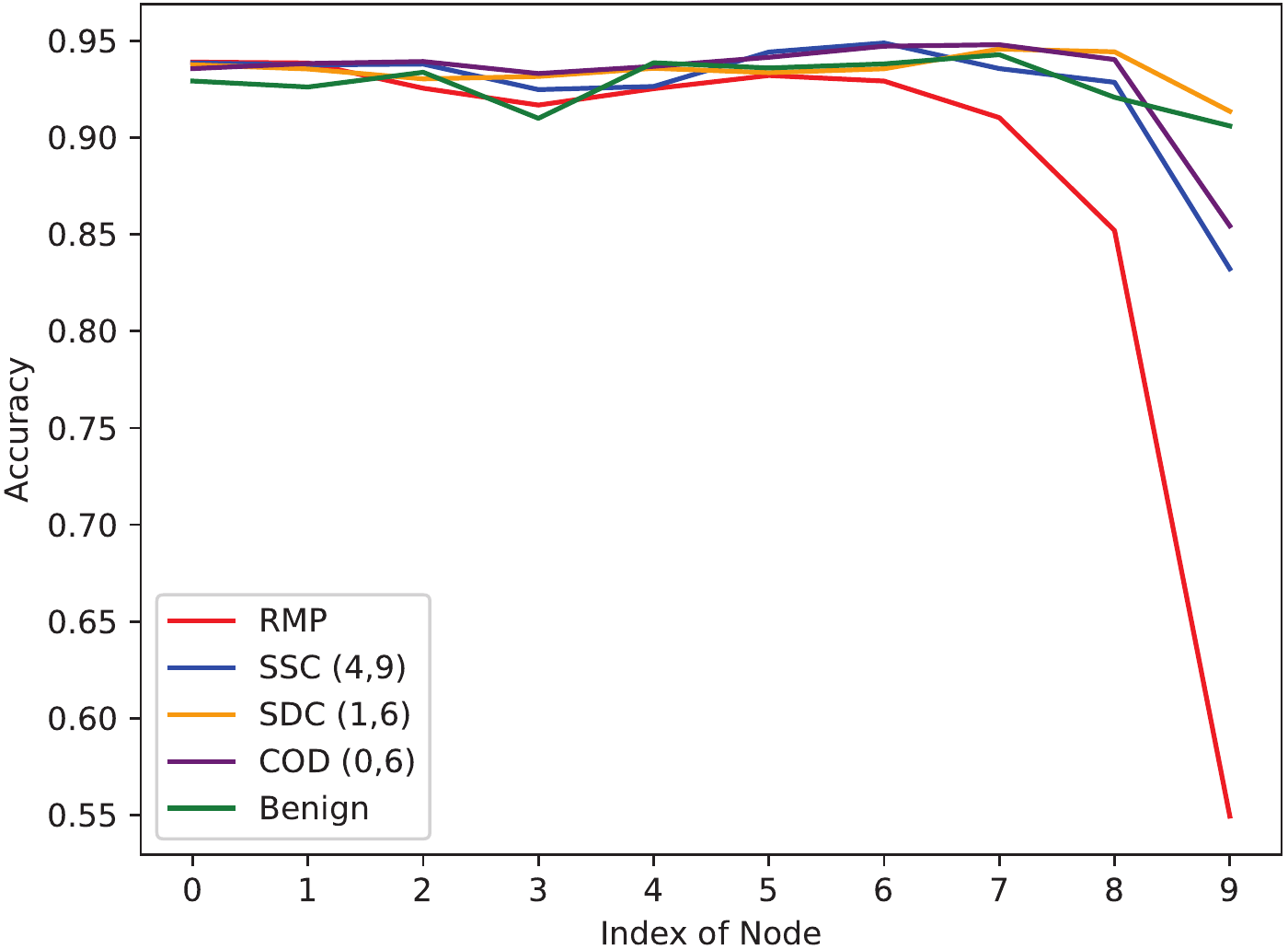}\\
  \scriptsize{\hspace{5em}(e) Under the Attack on Node 9}
  \end{minipage}
  \begin{minipage}[b]{0.32\textwidth}
  \raggedright
  \includegraphics[keepaspectratio, scale=0.38]{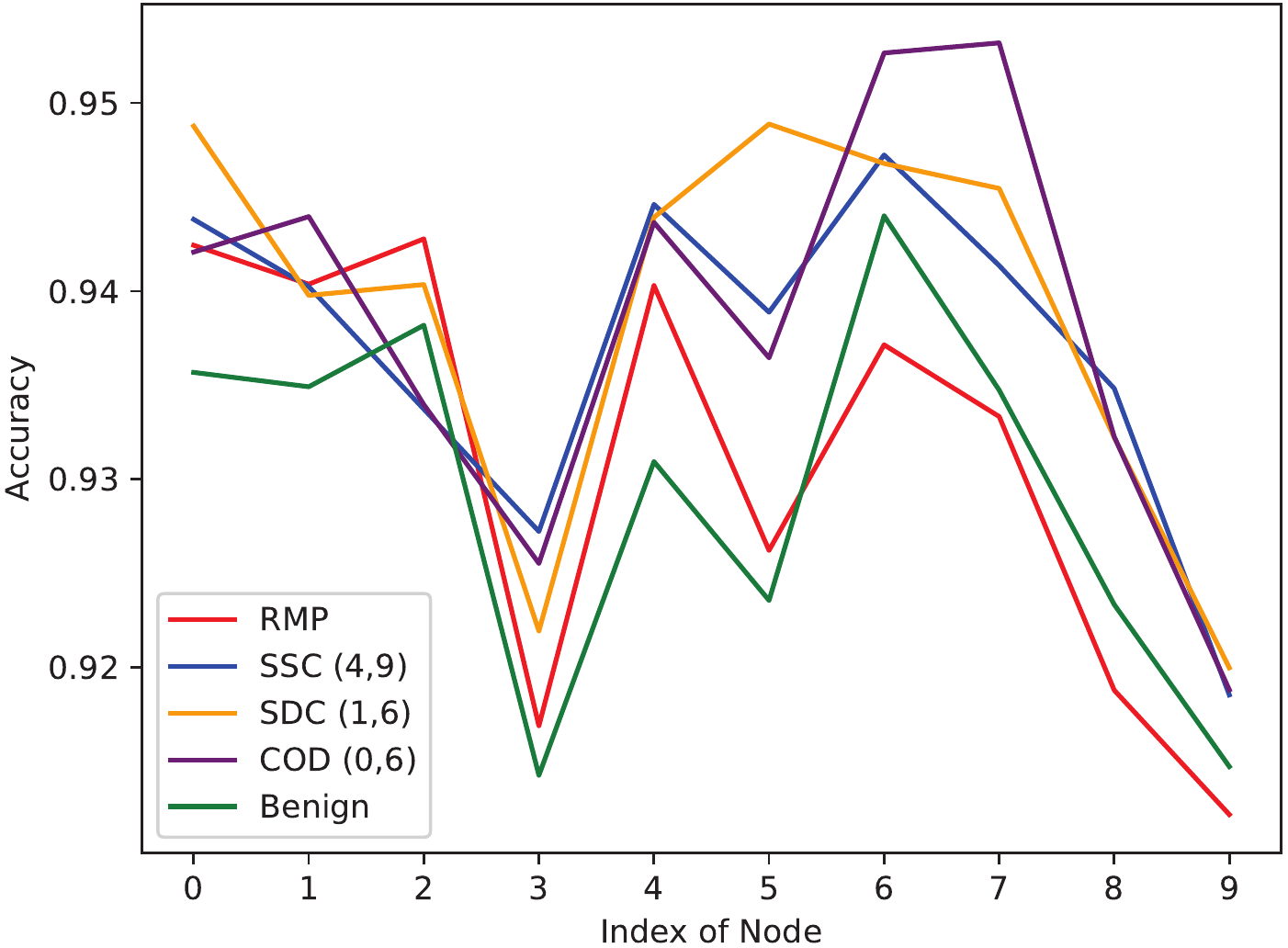}\\
  \scriptsize{\hspace{5em}(f) After the Attack on Node 9}
  \end{minipage}

\caption{Accuracy before, under, and after attacks by the node. In (a)-(c), the attacker is mounted on Node 0, whereas in (d)-(f), the attacker is on Node 9. Under the attack, the nodes next to the attacker suffered from degraded accuracy but other nodes were almost similar to the Benign cases. After the attack, i.e., under the recovery, attack-experienced models achieved higher accuracy overall compared to the Benign case.}
\label{fig:PoisonImpactAndRecovery}
\end{figure*}

\section{Evaluation}

We have carried out experiments to evaluate the resilience of WAFL against model poisoning attacks. To see the propagation of poisons and their resilience, we focus on the static line topology. The evaluation includes the analysis of the impact on accuracy, distortions of confusion matrices, and model disruptions and recovery. 

\subsection{Experiment Setting}

\subsubsection{Non-IID Dataset}

As WAFL assumes a Non-IID scenario, we have also configured
10 training nodes (n = 0, 1, . . . , 9) to have 90\% Non-IID
MNIST dataset. This is generated so that 90\% of node n’s data is composed of
label n’s samples and the other 10\% is uniformly composed
of the other label’s samples without any overlaps among the
nodes. Table \ref{tab:noniid_distribution} describes the detail of data distributions. We use the Non-IID dataset for training, but the standard MNIST test dataset, which is IID, for testing. 

\subsubsection{Machine Learning Model}

Based on the work\cite{ochiai2022wireless}, We have used a 2-layer fully connected neural network (FC-NN), which is configured as FC ($in=784, out=128$) - ReLU - FC ($in=128, out=10$). We have given random values as the initial model parameters before any training. We have used CrossEntropy as a loss function and Adam as an optimizer. We have used batch size 64, learning rate $\eta=0.001$, and the coefficient $\lambda=0.1$.

\subsubsection{Attack Scenario}

To observe the poisoned model propagation, we have configured a static network scenario named static\_line by \cite{ochiai2022wireless}. This network is statically connected as shown in Fig. \ref{fig:resilience_overview}. 
We assumed legitimate training from epoch 0 to 500, and then an attacker joins at Node 0 or at Node 9, persistently pushing constant poisoned models at each epoch. The attacker leaves the network at epoch 2000, and the whole training process ends at epoch 2500.

The poisoned models we prepared are RMP, SSC (4,9), SDC (1,6), and COD (0,6). RMP is generated by just initializing the model's network at random. SSC (4,9) is generated by training with the MNIST training dataset by swapping the labels between 4 and 9. In the same way, SDC (1,6) by swapping 1 and 6. COD (0,6) is generated by reassigning label 6 to label 0 data samples. We have run 100 epochs with a batch size of 64, a learning rate of 0.0001, and the Adam optimizer.

In the COD attack, we have specifically considered depressing label 0 because we specifically evaluate the attack from node 9. Because of the Non-IID configuration, only node 0 has labeled 0 samples primarily and strongly contributes to updating the model to classify and output 0. With this configuration, we observe the force balance between attacker and node 0 -- node 0 defends the label 0 output, whereas the attacker mounted at node 9 depresses the output of 0. 


  
  

\subsection{Confusions under Attack}

Fig. \ref{fig:ConfusionMatrix} shows the confusion matrices of Nodes 4, 6, 9, and Attacker under the attacks at epoch 2000. In this configuration, the attacker is mounted at Node 9. Those results show that Node 9 can suffer from the poisoned model but other nodes did not suffer from it, indicating great resilience to the poisoning attack.

\subsubsection{Under RMP Attack} 

We can observe that Node 9 suffered from the poisoned model from the attacker, whereas Node 6 did not suffer so much. In node 6, we can observe some misclassification as class 6 but this is simply because of the Non-IID feature, i.e., Node 6 has a much larger amount of class 6 data. This is also the same as Node 4. Especially, in node 4, some class 9 samples are predicted as 4 because classes 4 and 9 are potentially similar in the feature space. This is not because of the RMP attack.

\subsubsection{Under SSC (4,9) Attack}

We can observe that Node 9 could not predict class 4 samples correctly but predicted them as class 9 as the attacker intended. However, Node 9 could predict class 9 almost successfully showing its resilience to the attacker. We can observe other misclassifications in classes 7 and 8 at Node 9. We could not observe any clear effects of the poisoned model at Node 6. However, at Node 4, we can also observe a small increase in the misclassification of class 9 samples. This might be the effect of the poisoned model propagated to Node 4. 

\subsubsection{Under SDC (1,6) Attack}

We can observe a small increase in misclassification in classes 1 and 6 at Node 9. But, this increase was much smaller than SSC (4,6)'s case. Node 9 showed good resilience to the attacker's poisoned model. We could not observe any impacts of the poisoned model at Nodes 4 and 6 as discussed in RMP Attack result observations.

\subsubsection{Under COD (0,6) Attack}

We can observe that Node 9 suffered from outputting class 0, as the attacker intended. The class 0 samples were misclassified into classes 6, 9, 7, 5, and so on. The output of class 0 at Node 6 seems slightly depressed. There seems no influence on Node 4 as discussed in RMP Attack result observations.

\subsection{Accuracy - Before, Under, and After Attack }




\begin{figure*}
\centering

  \begin{minipage}[b]{0.31\textwidth}
  \raggedright
  \includegraphics[keepaspectratio, scale=0.38]{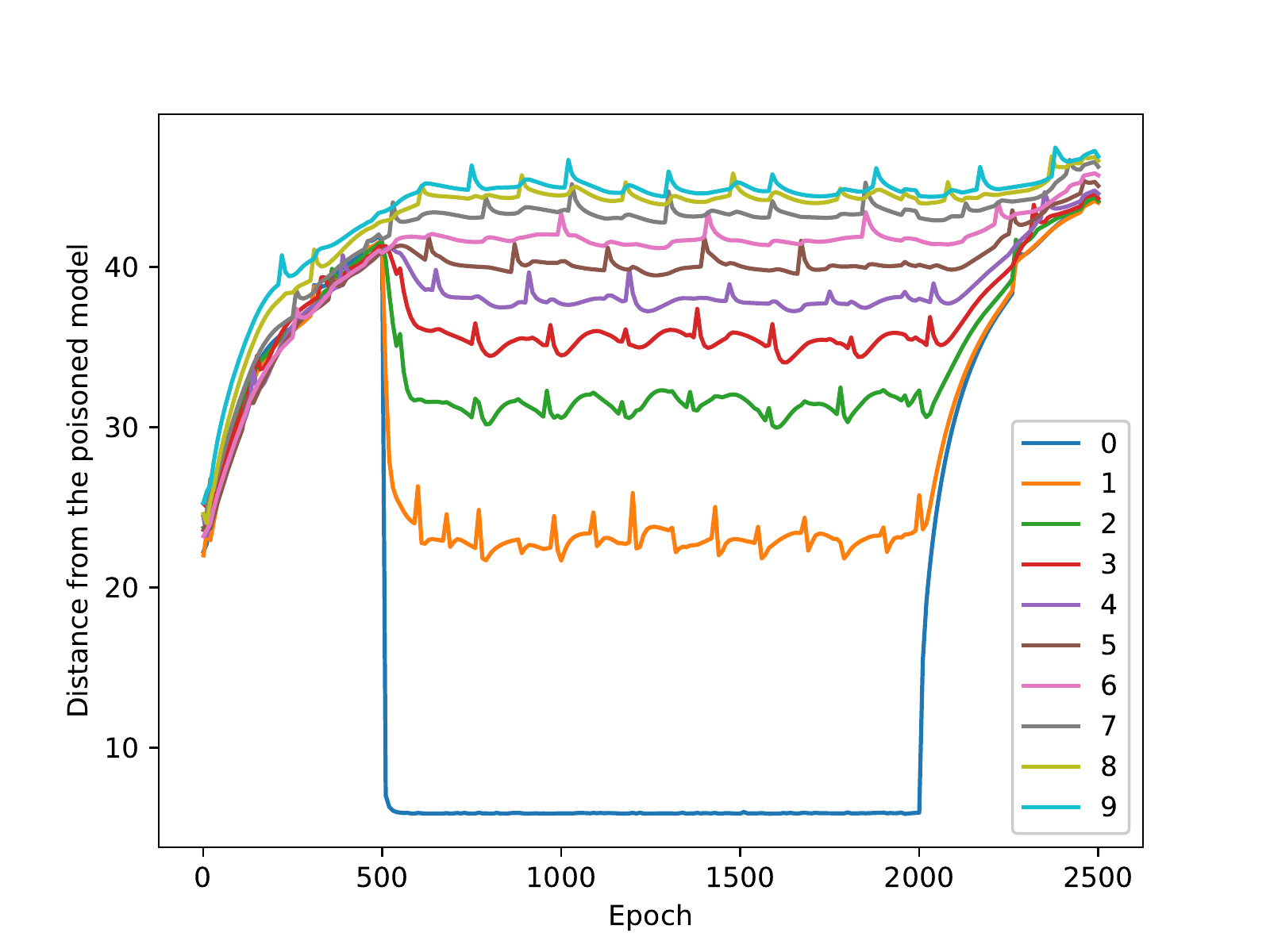}\\
  \scriptsize{\hspace{6em}(a) RMP Attack from Node 0}
  \end{minipage} 
  \begin{minipage}[b]{0.31\textwidth}
  \raggedright
  \includegraphics[keepaspectratio, scale=0.38]{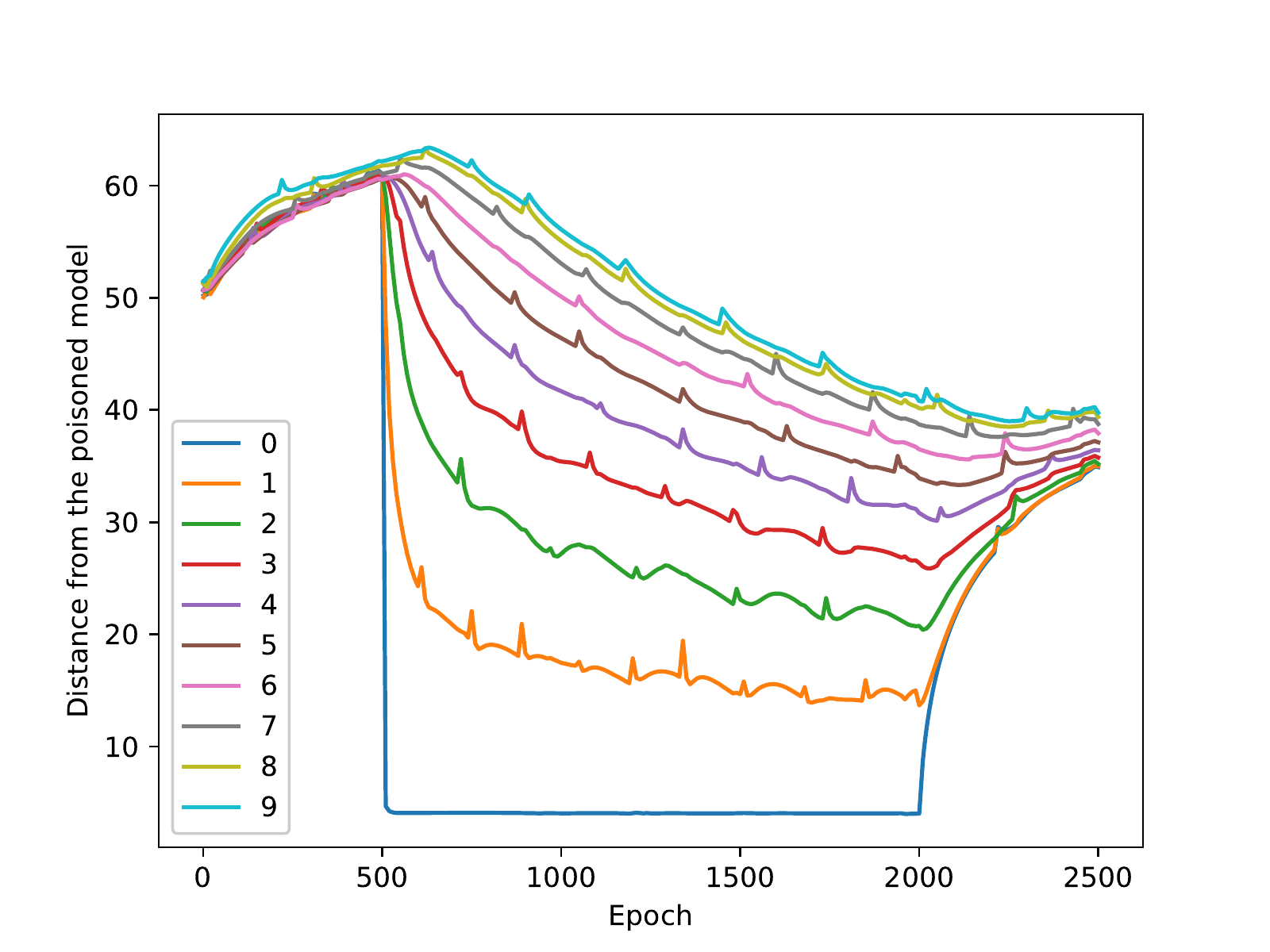}\\
  \scriptsize{\hspace{5em}(b) SSC (4,9) Attack from Node 0}
  \end{minipage}  
  \begin{minipage}[b]{0.31\textwidth}
  \raggedright
  \includegraphics[keepaspectratio, scale=0.38]{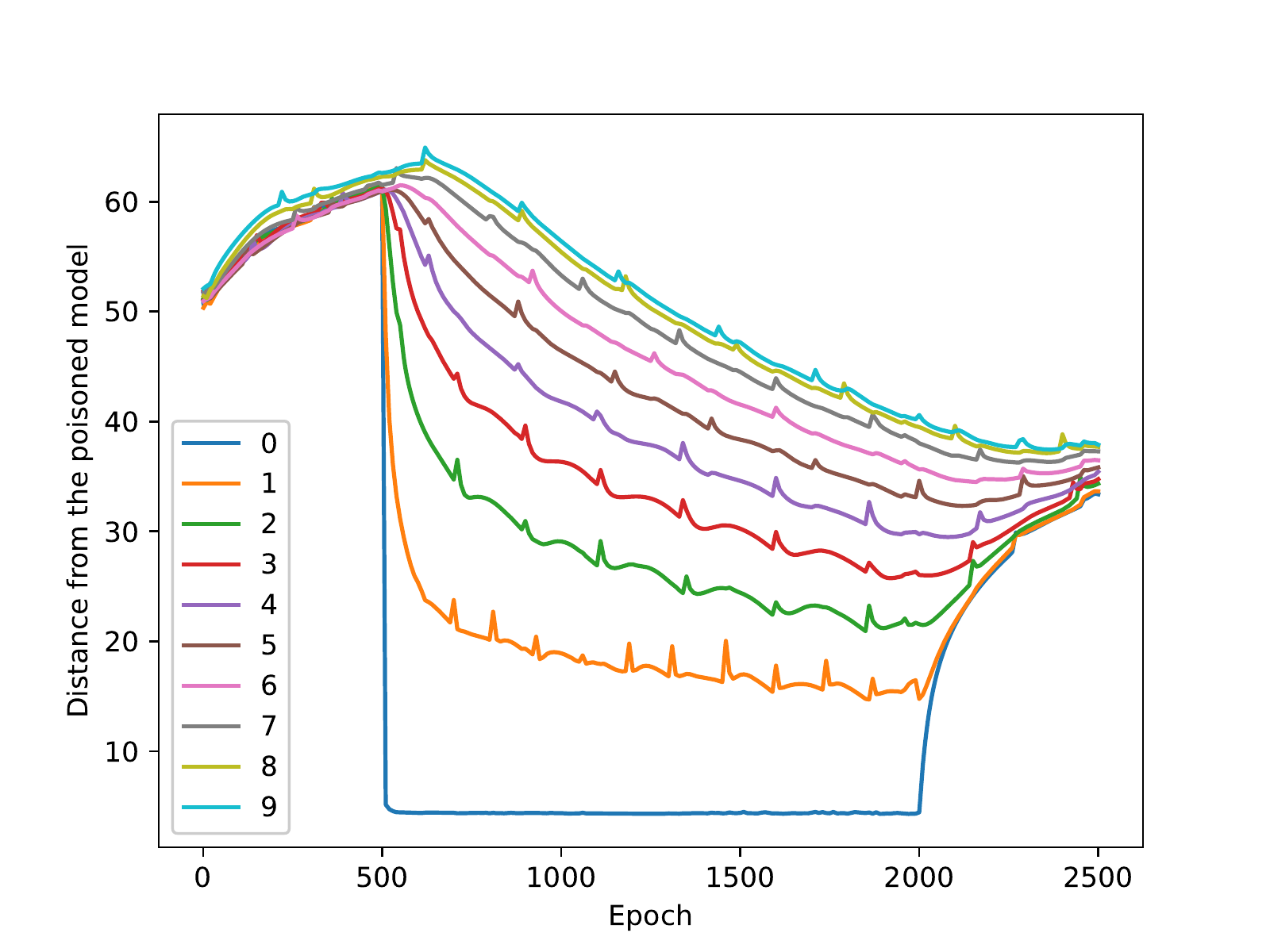}\\
  \scriptsize{\hspace{5em}(c) SDC (1,6) Attack from Node 0}
  \end{minipage}

  \begin{minipage}[b]{0.31\textwidth}
  \raggedright
  \includegraphics[keepaspectratio, scale=0.38]{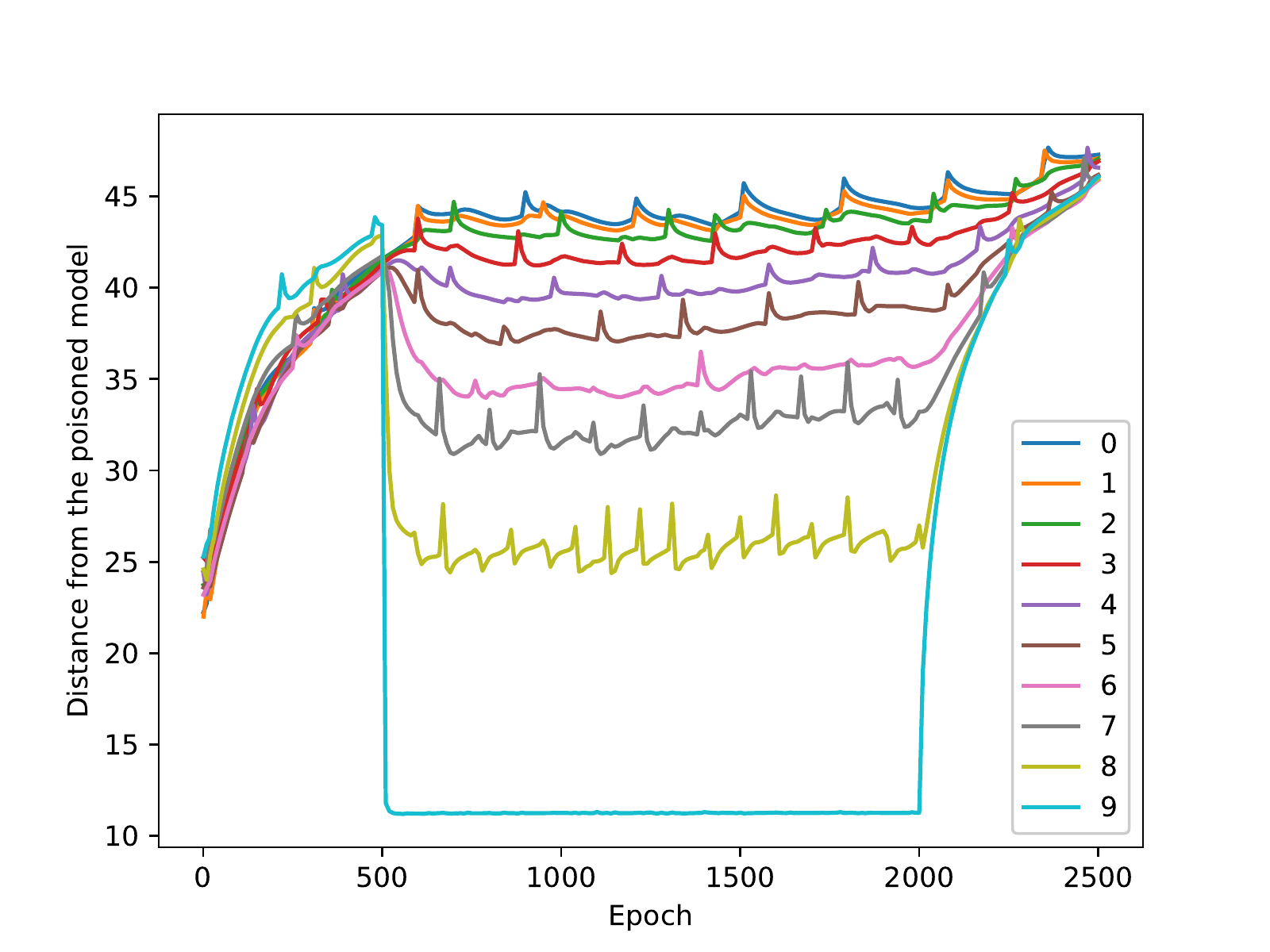}\\
  \scriptsize{\hspace{6em}(d) RMP Attack from Node 9}
  \end{minipage} 
  \begin{minipage}[b]{0.31\textwidth}
  \raggedright
  \includegraphics[keepaspectratio, scale=0.38]{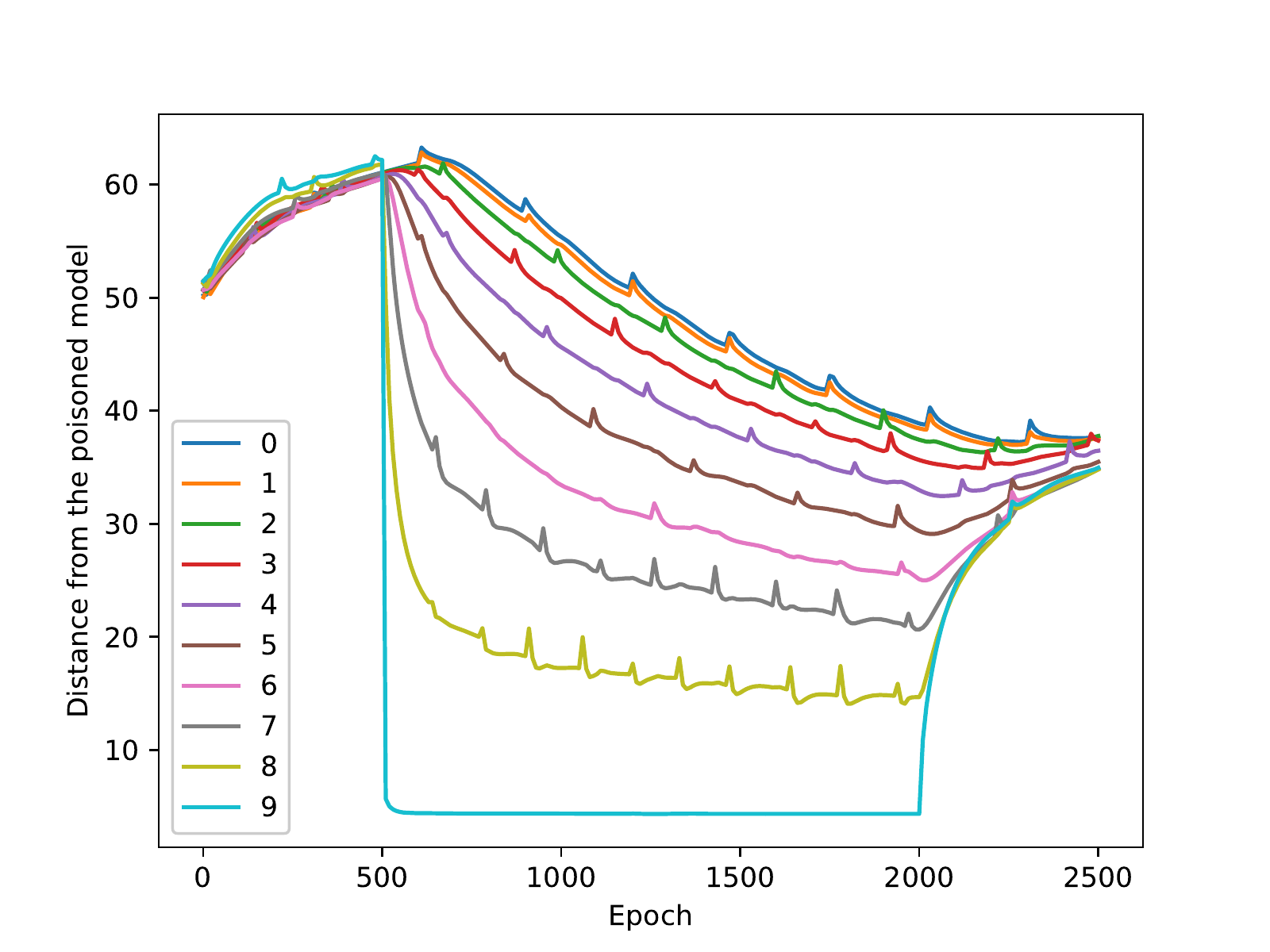}\\
  \scriptsize{\hspace{5em}(e) SSC (4,9) Attack from Node 9}
  \end{minipage}  
  \begin{minipage}[b]{0.31\textwidth}
  \raggedright
  \includegraphics[keepaspectratio, scale=0.38]{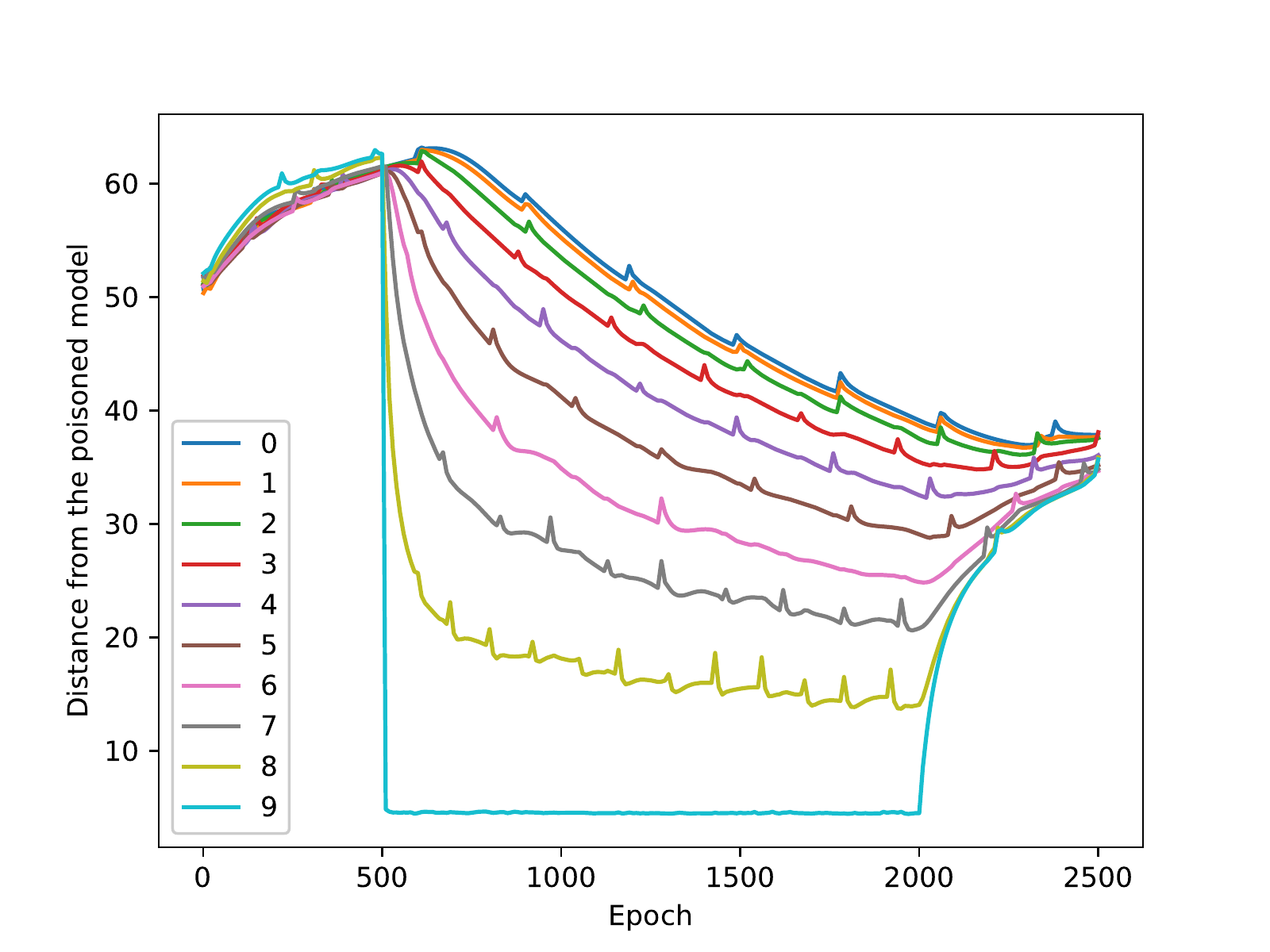}\\
  \scriptsize{\hspace{5em}(f) SDC (1,6) Attack from Node 9}
  \end{minipage}
  
\caption{Trend of model parameter distance from the poisoned model of the attacker on "fc1.weight". The color indicates the node number. In (a)-(c), the attacker is mounted on Node 0. In (d)-(f), the attacker is mounted on Node 9. The attack duration was from epoch 500 to 2000. The model parameters were driven to the poisoned model but showed resilience, especially in the case of the RMP attack. In the case of SSC and SDC, all the model parameters were shifted to the poisoned model, but the accuracy did not drop as Fig. \ref{fig:PoisonImpactAndRecovery} indicated.}
\label{fig:ModelDistance}
\end{figure*}

Fig \ref{fig:PoisonImpactAndRecovery} shows the accuracy of models per node before the attack, under the attack, and after the attack, i.e., under the recovery. In 
Fig \ref{fig:PoisonImpactAndRecovery} (a)-(c), the attacker was mounted on Node 0, whereas in (d)-(f), the attacker was mounted on Node 9.

To obtain averaged accuracies, we have calculated the averages of epochs 450 to 500 as ``Before the Attack'', epochs 1950 to 2000 as ``Under the Attack'', and epochs 2450 to 2500 as ``After the Attack''. We have taken this approach because those values sometimes have sudden changes because of the Adam optimizer.

Before the attack, the models of the nodes achieved accuracies around 0.88 and 0.92. These accuracies between Attacks and Benign were the same. Under the attack, there were accuracy drops at the next node or the next two-three nodes of the attacker. However, other nodes achieved more than 0.90 sometimes showing accuracy increases before the attacks. These results indicate the great resilience to the model poisoning attacks even under the attack.

Finally, after the attack, under the recovery phase, the nodes achieved accuracy between 0.91 and 0.95, which is higher than before the attack. 

More importantly, we have found that attack-experienced cases achieved higher accuracies than the Benign case in Fig. \ref{fig:PoisonImpactAndRecovery} (c) and (f) against the intention of the attacker. The poisons might have contributed to finding a better optimal model by the disturbance. In this way, WAFL has shown strong resilience in accuracy against model poisoning attacks.

\subsection{Distance from the Poisoned Model}

To track the change of model parameters, before, under, and after the attack. We evaluated the distance of the nodes' model from the poisoned model. For this, we have defined the following formula as a metric of distance from the poisoned model,
\begin{equation}
Distance(\theta^{(n)}) = \sqrt{\sum_{i=1}^{\vert \theta \vert} (\theta^{(n)}[i]-\theta^{(m)}[i])^2}.
\end{equation}

Here, we can consider $\theta$ as a partition of the whole model parameters. Specifically, we have picked up ``fc1.weight'' for visualizing the distance from the poisoned model. Here, fc1 stands for the first fully-connected layer which has 784 inputs and 128 outputs as indicated in Section IV.A.(2).

Fig. \ref{fig:ModelDistance} shows the result. We have only presented RMP, SSC (4,9), and SDC (1,6) in this example because COD (0,6) has shown a similar trend as SSC and SDC. The color of the graph indicates the node number.

In all the cases, model parameters were shifted toward the poisoned model during the attack. The hop counts corresponded to the model distance from the poisoned model as Fig. \ref{fig:parameter_space}. 

We observed two types of resilience in these results. 

\subsubsection{Resilience in Model Parameter}
In the case of RMP, The distances were almost stable during the attack. This indicates the resilience of the model parameter itself from the poisoned one. 

\subsubsection{Resilience in Accuracy}
In the case of SSC and SDC, all the distances were decreased to almost half. However, this does not mean a drop in accuracy as discussed in Section IV.C. This indicates the tolerance to model poisoning. Even if poisoned model parameters are injected and remain there, the model can change itself to maintain or even achieve higher accuracy as discussed with Fig. \ref{fig:PoisonImpactAndRecovery} (c) and (f).

\section{Future Research Directions}

We have studied the resilience of WAFL against model poisoning attacks, providing an analysis of the propagation of the poisoned model. In this paper, we focused on the static\_line network topology and MNIST dataset classification problem in the Non-IID scenario as a benchmark evaluation. As a result, we have found that WAFL can show good resilience in both model parameters and accuracy against model poisoning attacks. We have even found that attack-experienced models can achieve higher accuracy because the disturbance can contribute to finding a better optimal model. We here list further research items.

\subsection{Other Network Topology}
The application of WAFL\cite{ochiai2022wireless} is discussed with a wide variety of node-to-node networks including dynamically-changing opportunistic networks such as random waypoint mobility\cite{camp2002survey} and community-structured environment\cite{ochiai2008mobility}. The analysis of poisoned model propagation and resilience on these networks would be one of the future research directions.

\subsection{Backdoor-enabled Model Poisoning}

As discussed in related work, Backdoor-enabled models\cite{li2022backdoor} would be another important poisoning model related to our study. Because WAFL has great resilience to the poisoned models injected, such backdoor-enabled poisoned models might be disrupted during the propagation, losing their functionality as a backdoor, which should be further studied in the future.

\subsection{Detection of Model Poisoning}

Detection mechanisms can be also studied in future work. For example, from Fig. \ref{fig:ModelDistance}, we can expect that a legitimate node might be able to detect a poisoning attack by checking the distances of the received models. However, we will need to carefully study this as future work, because some persistent attack could slowly change the poisons, or in dynamic networks, a node may encounter a largely-different legitimate model. 

\section{Conclusion}

In this paper, we have presented our analysis of the WAFL's resilience against model poisoning attacks. As WAFL is fully distributed and collaboratively trains generalized models, WAFL shows great resilience to a poisoned model injected by an attacker. This force balance has been theoretically analyzed in this paper. Our experiment results have shown that the nodes directly encountered the attacker has been somehow compromised to the poisoned model but other nodes have shown their resilience.  All the nodes have finally found stronger model parameters combined with the poisoned model. Consequently, most of the attack-experienced cases achieved higher accuracy than the no-attack experienced cases after the recovery.

\section*{Acknowledgment}
This work was supported by JSPS KAKENHI Grant Number JP 22H03572.


\bibliographystyle{unsrt}
\bibliography{waflattack.bib}

\end{document}